\documentclass[conference]{IEEEtran}
\IEEEoverridecommandlockouts
\usepackage{graphicx} 
\usepackage{pifont}
\usepackage{amsmath}
\usepackage{amssymb}  
\usepackage{amsfonts}
\usepackage{algorithmic}
\usepackage{array}
\usepackage{multicol}

\usepackage{textcomp}
\usepackage{stfloats}
\usepackage{url}
\usepackage{verbatim}
\usepackage{graphicx}
\usepackage{caption}
\usepackage{subcaption}
\usepackage[ruled, vlined, linesnumbered]{algorithm2e}
\usepackage{cite}
\usepackage{booktabs}
\usepackage{amssymb,amsfonts,bm}
\usepackage{amsmath,tikz}
\usepackage{color}
\usepackage{mathtools}
\usepackage[hidelinks]{hyperref} 
\usepackage{rotating}
\usepackage{blkarray}
\usepackage{physics}
\usetikzlibrary{arrows}
\usepackage{makecell}

\usepackage{amsthm}
\usepackage{comment}
\usepackage{multirow}
\usepackage{geometry}
\graphicspath{{images/}}

\begin{document}

\title{\LARGE \bf
MMAUD: A Comprehensive Multi-Modal Anti-UAV Dataset for Modern Miniature Drone Threats\\

\author{Shenghai Yuan$\dagger$, ~Yizhuo Yang$\dagger$,~Thien Hoang Nguyen, ~Thien-Minh Nguyen, ~Jianfei Yang,\\ Fen Liu,~Jianping Li,~Han Wang,~Lihua Xie*,~\textit{Fellow,~IEEE}

\thanks{This research is supported by the National Research Foundation, Singapore, under its Medium-Sized Center for Advanced Robotics Technology Innovation (CARTIN) and Delta-NTU Corporate Lab. }
\thanks{$\dagger$ Equal contribution. All authors are with the School of Electrical and Electronic Engineering, Nanyang Technological University, 50 Nanyang Avenue, Singapore 639798, 
   { Email: \tt\small  \{shyuan, thien.nguyen, thienminh.nguyen, jianfei.yang, fen.liu, jianping.li, elhxie\} @ntu.edu.sg, \{yang0670, hwang027\}@e.ntu.edu.sg.}}%

}

}

\maketitle

\begin{abstract}
In response to the evolving challenges posed by small unmanned aerial vehicles (UAVs), which possess the potential to transport harmful payloads or independently cause damage, we introduce MMAUD: a comprehensive Multi-Modal Anti-UAV Dataset. MMAUD addresses a critical gap in contemporary threat detection methodologies by focusing on drone detection, UAV-type classification, and trajectory estimation.
MMAUD stands out by combining diverse sensory inputs, including stereo vision, various Lidars, Radars, and audio arrays. It offers a unique overhead aerial detection vital for addressing real-world scenarios with higher fidelity than datasets captured on specific vantage points using thermal and RGB. Additionally, MMAUD provides accurate Leica-generated ground truth data, enhancing credibility and enabling confident refinement of algorithms and models, which has never been seen in other datasets.
Most existing works do not disclose their datasets, making MMAUD an invaluable resource for developing accurate and efficient solutions. Our proposed modalities are cost-effective and highly adaptable, allowing users to experiment and implement new UAV threat detection tools.
Our dataset closely simulates real-world scenarios by incorporating ambient heavy machinery sounds. This approach enhances the dataset's applicability, capturing the exact challenges faced during proximate vehicular operations.
It is expected that MMAUD can play a pivotal role in advancing UAV threat detection, classification, trajectory estimation capabilities, and beyond. Our dataset, codes, and designs will be available in \href{https://github.com/ntu-aris/MMAUD}{https://github.com/ntu-aris/MMAUD}.
\end{abstract}

\begin{IEEEkeywords}
UAV, LIDAR, Audio, video fusion, Detection, Classification, Trajectory Estimation.
\end{IEEEkeywords}

\section{Introduction}
In an era where compact commercial drones, easily accessible as commercial off-the-shelf products (COTS), exhibit remarkable capabilities, their potential for misuse looms greatly. These drones boast extensive ranges, the capacity to operate at high altitudes, and the ability to minimize heat and acoustic emissions, rendering them inconspicuous and ideal for unauthorized access to restricted areas or repurposing for potentially harmful activities.

Recent war conflicts have highlighted the adaptability of these civilian drones, which successfully evaded even portable air defense systems reliant on infrared or Radar targeting. Their capacity to reduce sensor signatures challenges conventional detection methods, presenting a cost-effective means of achieving aerial stealth and becoming a threat to lives.

\begin{figure}[t]
\centering
\includegraphics[width=8.6cm]{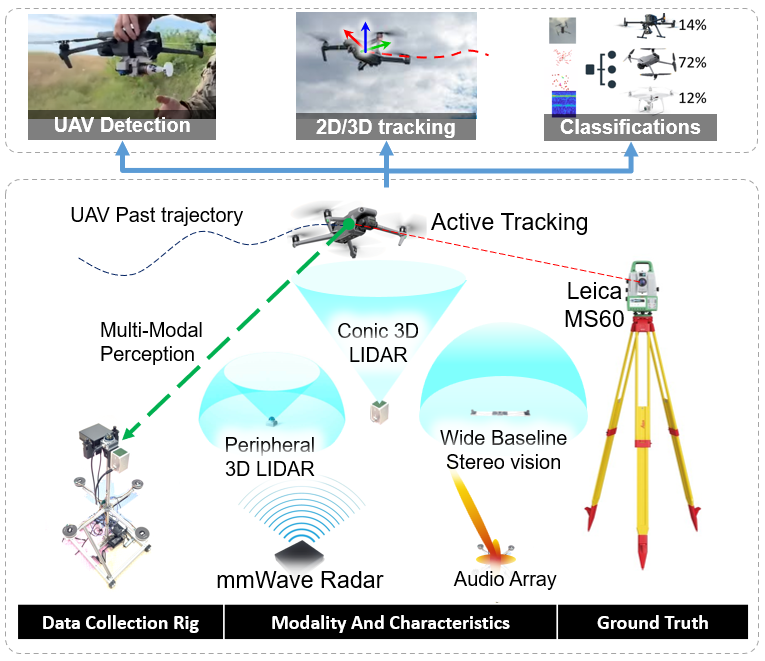}
\caption{System Overview.}
\label{fig:system_overview}
\vspace{-2em}
\end{figure}

To address these evolving challenges, we present our Anti-UAV dataset, a comprehensive dataset for detecting, classifying, tracking, and estimating the trajectories of such drones. Our contributions can be summarized as follows:

\begin{enumerate}
  \item We introduce a multi-modal dataset that integrates visual, LIDAR array, RADAR, and audio array sensors, offering a rich and diverse data source for advanced UAV detection techniques as shown in Fig. \ref{fig:system_overview}.
  \item 
Utilizing ground truth data generated by Leica, our dataset sets unprecedented millimeters level accuracy benchmarks in the field of anti-UAV datasets, a distinctive feature not present in previous datasets.
  \item Our approach embraces cost-effective sensor configurations and open-source code, facilitating the development of mobile-ready, life-saving applications. The code and dataset are openly accessible to the community, encouraging collaborative research and innovation.
\end{enumerate}

\vspace{-0.25em}
\section{Related Work}
To the best of our knowledge, there are few datasets that encompass multi-modal perception tools for tasks such as detection, tracking, classification, and trajectory estimation of COTS UAVs, as shown in Table \ref{tab:datasetcompare}. Most community datasets for tracking primarily focus on common objects like cars \cite{yu2020unmanned} and people \cite{yang2023mm}, with the potential to include new classes such as UAVs. Some 2D detection datasets provide images but lack 3D position ground truth and sufficient labeling. In recent years, anti-UAV challenges  \cite{jiang2021anti,zhao21082nd,zhao20233rd}, have emerged in CVPR. However, these datasets \cite{jiang2021anti},\cite{zhao21082nd},\cite{zhao20233rd},\cite{svanstrom2021dataset},\cite{zhao2022vision},\cite{chen2017deep} depend on costly thermal and RGB cameras strategically positioned at elevated vantage points, often requiring human intervention for precise aiming. Their primary focus is on object detection and 2D tracking, lacking the inclusion of 3D trajectory estimation. These datasets are primarily designed for computer vision-related challenges and are rarely suitable for real-world field applications.

\begin{table*}[t]

\centering
\caption{List of Public Available Drone Detection Dataset.}
\label{tab:datasetcompare}
\renewcommand{\arraystretch}{1.5}

\begin{tabular}{lcccccccccc}

\toprule 
\hline

\multirow{2}{*}{Dataset} & \multicolumn{5}{c}{Modality} & \multirow{2}{*}{Ground Truth} & \multicolumn{4}{c}{Task} \\
\cmidrule(lr){2-6} \cmidrule(lr){8-11}
& Imaging & Audio & Radar & LIDAR & RF & & Detection & Tracking & Classification & Trajectory \\
\midrule

    Anti-Drone\cite{jiang2021anti},\cite{zhao21082nd},\cite{zhao20233rd} & $\checkmark$ &  $\cross$ &  $\cross$ &  $\cross$ &  $\cross$ &  Manual  &  $\checkmark$ & $\checkmark$ &  $\cross$ &  $\cross$ \\ 

\hline

    Svanstr{\"o}m et al.\cite{svanstrom2021dataset} & $\checkmark$ &  $\checkmark$ &  $\cross$ &  $\cross$ &  $\cross$ &  Manual  &  $\checkmark$ & $\cross$ &  $\cross$ &  $\cross$ \\

\hline

    DUT-Anti-UAV\cite{zhao2022vision}& $\checkmark$ &  $\cross$ &  $\cross$ &  $\cross$ &  $\cross$ &  Manual  &  $\checkmark$ & $\checkmark$ &  $\cross$ &  $\cross$ \\

\hline

    Chen et al.\cite{chen2017deep}& $\checkmark$ &  $\cross$ &  $\cross$ &  $\cross$ &  $\cross$ &  Semi-Auto  &  $\checkmark$ & $\checkmark$ &  $\cross$ &  $\cross$ \\

\hline

    Wang et al.\cite{wang2022large}& $\cross$ &  $\checkmark$ &  $\cross$ &  $\cross$ &  $\cross$ &  Manual  &  $\cross$ & $\cross$ &  $\checkmark$ &  $\cross$ \\
\hline

    Zhang et al.\cite{zhang2023mmhawkeye}& $\checkmark$ &  $\checkmark$ &  $\cross$ &  $\cross$ &  $\checkmark$ &  Manual  &  $\checkmark$ & $\cross$ &  $\cross$ &  $\cross$ \\
\hline

    mDrone\cite{zhao20213d} & $\cross$ &  $\cross$ &  $\checkmark$ &  $\cross$ &  $\cross$ &  Full-Auto  &  $\checkmark$ & $\checkmark$ &  $\cross$ &  $\cross$ \\

\hline

    \textbf{MMAUD (Proposed)} & $\checkmark$ &  $\checkmark$ &  $\checkmark$ &  $\checkmark$ &  $\cross$ &  \textbf{Full-Auto} &  $\checkmark$ & $\checkmark$ &  $\checkmark$ &  $\checkmark$ \\
    \hline
    \bottomrule
\end{tabular}
\vspace{-2em}
\end{table*}

Certain studies have delved into the development of countermeasure systems against drones. Nevertheless, even among those that provide datasets\cite{zhao20213d}, they often provide limited modality and ground truth, primarily designed for indoor applications.  Furthermore, majority of other datasets \cite{svanstrom2021dataset,zhao2022vision,chen2017deep,zhang2023mmhawkeye,zhao20213d} utilize one or two modalities with manual or semi-automatic annotations, which poses challenges for comprehensive and generalizable evaluations. 
Some datasets \cite{wang2022large} focus on UAV classification through audio input, yet they lack spatial information and threat indication for the sensor suit. 

A new idea has been proposed by Zheng et al. \cite{zheng2022westlakedetection}, employing multi-view stereo setups for detecting and tracking nearby drones. However, their work is hindered by a lack of publicly available code and datasets. Furthermore, their proposed systems are costly and entail complex synchronization and computations of 16-32 camera sets. In light of recent war conflict experiences, the capability to identify overhead threats is essential, and affordable sensor configurations are favored to promote wider adoption for critical life-saving applications. 

In numerous cases, new methods\cite{shi2018anti,svanstrom2022drone,ding2023drone,bavu2022deeplomatics,hoffmann2016micro,farlik2016radar,christnacher2016optical,zhang2016intruder,ganti2016implementation,ghosh2023airtrack,mccoy2023ensemble}  for drone detection have been proposed, each claiming certain levels of accuracy and performance. Nonetheless, none of these approaches offer access to their datasets or source code, and there is a noticeable absence of discussion regarding the acquisition of precise ground truth, a facet that distinguishes our previous work\cite{nguyen2022ntu}. This absence of open datasets and code contributions raises concerns about the validity and reproducibility of their findings.

Other theoretical studies \cite{erskine2021model,liu2023non,xu2023shunted,nguyen2019integrated,cao2023neptune}, utilize range \cite{nguyen2019integrated,liu2023non}, or bearing\cite{erskine2021model}  observations for drone state estimation and tracking. However, these methods are generally regarded as impractical in real-world scenarios because they heavily depend on specific assumptions and can not be verified for effectiveness.

Radio Frequency (RF) detection is another method, but it faces challenges due to the wide frequency range of drones. Developing a single device for effective monitoring of all channels is difficult. Some drones don't emit RF signals, making RF detection unreliable.
Commercial drone detection systems like DJI Aeroscope\footnote{https://www.dji.com/sg/aeroscope} already exist, reducing the need for this modality in our dataset.

\vspace{-0.25em}
\section{Sensor Setup}
A custom-designed aluminum rig has been created to accommodate the entire sensor suite, as shown in Fig. \ref{fig:system_overview}. Table. \ref{tab:datasetcompare} presents a summary of the sensors along with their respective specifications. Each message is time-stamped based on its publication in ROS. Subsequently, we offer a more comprehensive overview of each sensor below.

\subsubsection{Stereo Cameras}
The MMAUD dataset includes two PIXELXYZ color cameras\footnote{https://www.pixelxyz.com/product/PXYZ-S-AR135-130T400.html}
 oriented upward, costing approximately 200 USD. These cameras are synchronized using an integrated trigger mechanism to ensure simultaneous image capture and transmission. Images are concatenated and transmitted through UVC camera protocols. With a baseline separation of approximately 17.8cm, the stereo setup enables robust stereo depth perception up to 20 meters, and the single camera can see a drone from 100 meters away in the ideal case at a resolution of 2560x960. Each camera provides an expansive 180-degree field of view, greatly enhancing the detection of UAVs throughout the horizon while forming a local dome-shaped detection volume. This notion of dome-shaped awareness bears a resemblance to earlier state-of-the-art research \cite{zheng2022westlakedetection}. However, our approach is significantly more cost-effective, utilizing a budget-friendly two-camera stereo setup instead of a 14k USD 32-camera stereo system. In addition to cost considerations, it's worth noting that despite being manufactured by a small workshop in China, these cameras deliver superior imaging quality compared to their Flir Blackfly counterparts. The ultimate goal is to transform this system into a wearable or mountable device for both humans and vehicles, allowing it to effectively mitigate UAV threats at a reasonable cost.

\subsubsection{Conic 3D LIDAR}
The proposed dataset also includes an upward-facing DJI Livox Avia\footnote{https://www.livoxtech.com/avia} LIDAR system designed for conic shape detection. This LIDAR system effectively covers a wide 70-degree conic field of view in the center and reliably detects objects at distances of up to 300 meters. Its scanning pattern is non-repetitive, which ensures the possibility of detecting drones within its field of view, although the presence of drones cannot be guaranteed consistently. While more cost-effective alternatives such as the DJI Livox Mid 50, Livox Mid-60, and Livox Horizon are available, this specific sensor was selected for internal IMU integration. This integration enables future use in moving vehicles, facilitating motion compensation for point cloud data. However, it's important to note that this feature will not be covered in this work.

\subsubsection{Peripheral 3D LIDAR}
In this research, we employ a peripheral horizontal placement of the DJI Livox Mid360\footnote{https://www.livoxtech.com/mid-360} LIDAR system, which proves to be a cost-effective solution. The Livox Mid360 offers an expansive 360° horizontal field of view and a 59° vertical field of view above the ground, making it highly capable of detecting objects within a range of up to 70 meters. Its non-repetitive scanning pattern ensures that all points within its field of view are effectively scanned, making it an ideal choice for detecting approaching rogue drones. This cost-efficient device plays a crucial role in enhancing the system's capability to detect nearby obstacles and potential threats.

\subsubsection{Audio Array}
One distinctive feature of drones is their unique noise emission. In our study, we've enhanced drone detection by incorporating four cost-effective Hikvision DS-VM1 omnidirectional microphone\footnote{https://www.hikvision.com/en/products/Turbo-HD-Products/Turbo-HD-Cameras/video-and-audio-conference-series/ds-uac-m1p/} arrays mounted on sensor rig, which is a low-cost 4-channel variant suitable for massive deployment. These arrays effectively capture human speech noise levels within a range of up to 10 meters. These microphones also excel at detecting louder drones at distances of 30-40 meters.
Remarkably, the four microphone arrays cost just 150 USD, making them budget-friendly and accessible for various applications.
We've strategically positioned the four microphones in a cross-shaped configuration, enabling the use of the precise differential time of arrival method to determine the direction and range of incoming drones effectively. This innovation enhances the ability to detect and respond to potential threats, making it a valuable asset in diverse scenarios.

\begin{figure}[t]
\centering
\includegraphics[width=8.6cm]{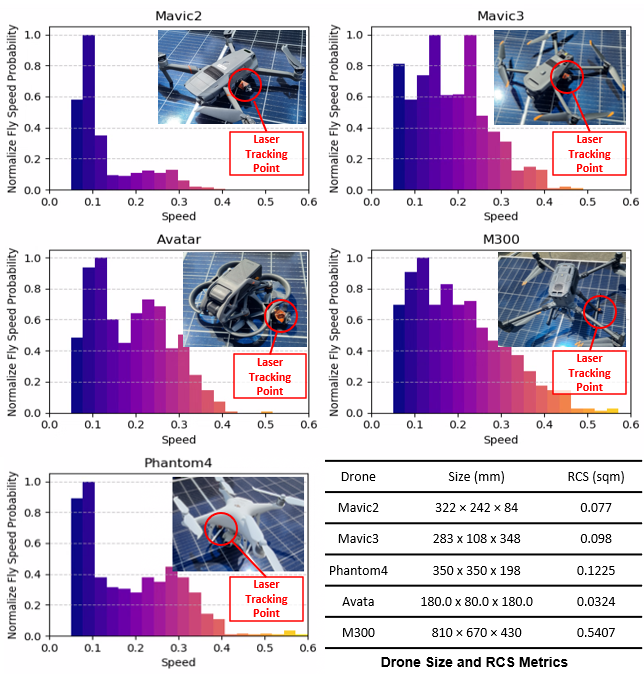}
\caption{List of UAV being collected and their characteristics. Notes: The RCS is approximated based on the size projection. }
\label{fig:drone_images}
\vspace{-1em}
\end{figure}

\subsubsection{Millimeter Wave Radar}
Lastly, we've introduced a high-value addition to our dataset - the Oculii Eagle ETH04 77GHz Millimeter wave point cloud imaging Radar 4D imaging radar\footnote{https://www.oculii.com/eagle}, with an approximate cost of 3.6k USD. This radar represents one of the most valuable components in our dataset, as it aligns with the systems used for UAV threat detection in some literature and real-world deployed systems.

Our selection of this unit is based on its remarkable attributes, notably an extended detection range and superior packaging compared to radars from Taxus Instruments (TI), such as the TI AWR1843\footnote{https://www.ti.com/product/AWR1843}. The Eagle ETH04 radar boasts a horizontal field of view spanning 120 degrees and a vertical field of view of 30 degrees. It excels in sensing moving objects at distances of up to 350 meters. This radar's enhanced capabilities make it a standout inclusion in our dataset, further enriching its potential applications and effectiveness in various scenarios.

\subsubsection{Ground Truth}
Utilizing a Leica Nova MS60 MultiStation\footnote{https://leica-geosystems.com/en-us/products/total-stations/multistation/leica-nova-ms60}, we track a crystal prism positioned on the UAV, which serves as a valuable source of ground truth for our position estimations, as illustrated in Fig. \ref{fig:data_visualization}. It's noteworthy that the coordinate frame adopted by this ground truth system aligns with the gravitational field during the startup process, resulting in its z-axis pointing opposite to the direction of gravity. 
The Leica tracking system recorded UAV ground truth locations at 5Hz.
To ensure accurate synchronization, we segmented the data based on the timestamps provided by the Leica system. This allowed us to precisely align each image with its corresponding ground truth location.

\section{Dataset Characteristics}
\begin{figure}[t]
\centering
\includegraphics[width=8.6cm]{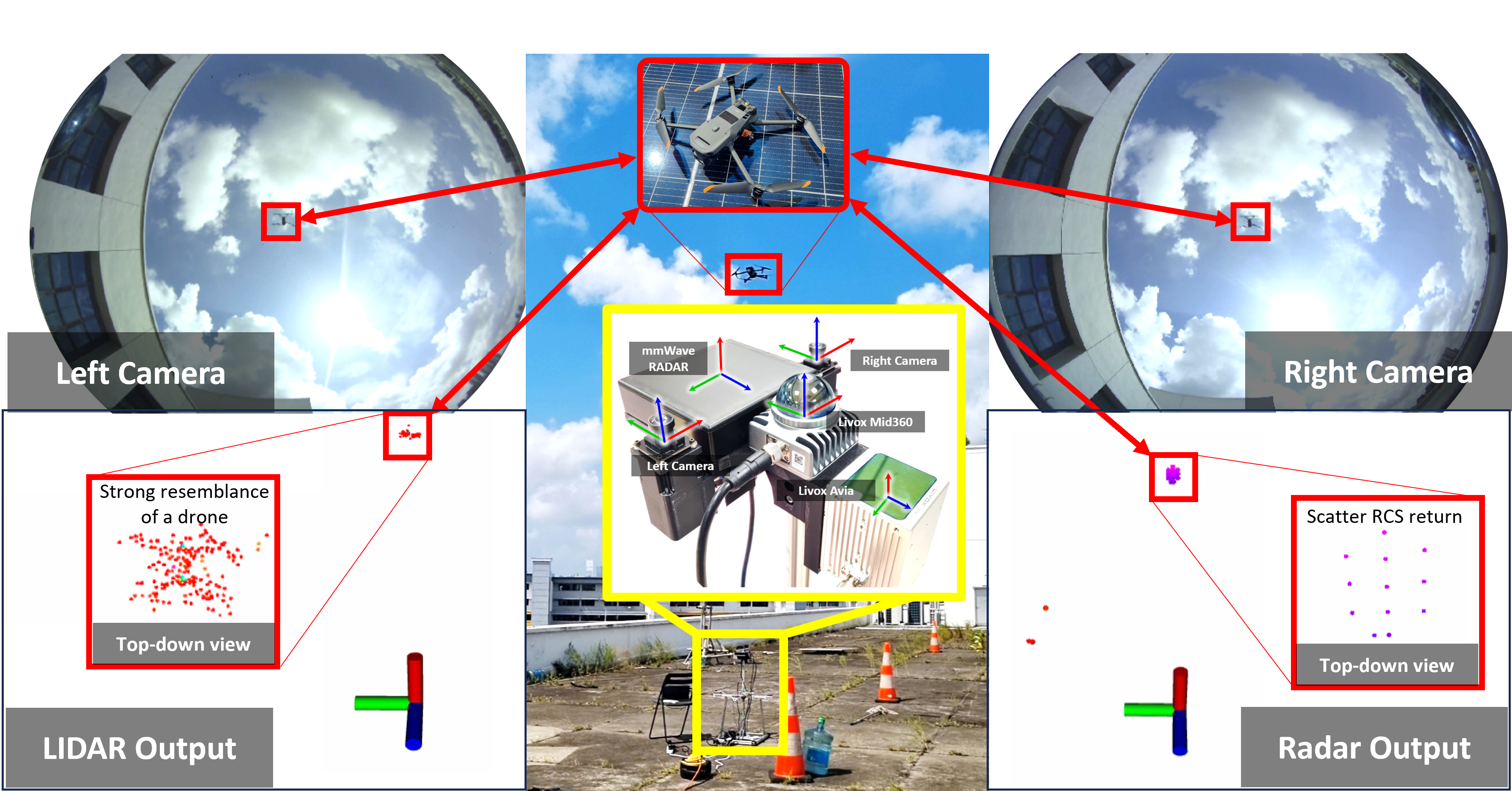}
\caption{Visualization of the Visual, LIDAR and RADAR data.}
\label{fig:data_visualization}
\vspace{-2em}
\end{figure}

The datasets are categorized into six groups, each corresponding to different drone types: Mavic2, Mavic3, Avata, Phantom4, M300, and ambient noise sequences, as shown in Fig. \ref{fig:drone_images}. 
The speed, size, and estimated RCS of the drone are also modeled accordingly with precision by the relative ground truth.
Each sequence encompasses visual, 4xaudio, 2xLIDAR, and RADAR information, as depicted in Fig. \ref{fig:data_visualization}.

Recognizing the importance of audio modality, we conducted tests in outdoor environments to create a more realistic setting based on the review from our previous work\cite{Yizhuo2023iros}. Specifically, we selected locations characterized by ambient noise, including the operation of heavy machinery and the presence of powerful air-conditioning systems. Additionally, at higher altitudes, wind noise added complexity to the audio data, making detection more challenging and enhancing the realism of our system.

It's worth noting that we did not include data collected during nighttime or rainy conditions. This decision is rooted in the fact that drones are easily detectable at night due to their lights, and they face operational challenges in rainy weather, making such data less relevant for our dataset.

\section{Dataset Format}

Our dataset is conveniently accessible in two widely used formats: the rosbag format and the filesystem format. The specific data structure within the rosbag format is shown in Table \ref{tab:sensorspecificaions}.
\\
\textbf{Rosbag format}: Rosbag is widely used in the robotics community. While all measurements adhere to SI units, it's essential to note that message timestamps utilize the ros::Time format, residing in the header.stamp field. These timestamps are generated by the sensor driver when ROS messages are published. Additionally, the LIDAR point cloud employs custom formats featuring individual timestamps for each point. We provide scripts to facilitate the conversion of these points into the standard sensor\_msgs::pointcloud2 format.\\
\textbf{Filesystem format}: 
The filesystem format is widely favored within the machine learning community. Essentially, it involves the process of extracting images, raw audio data, point clouds, and ground truths from the rosbag file and then storing them sequentially for convenient access and management.
To offer further clarification, this entails saving the images in PNG format, breaking down the audio data into 1-second segments stored in numpy format, preserving the point cloud data in PCD format, and retaining the 3D position data at each timestamp in a python-numpy file for future reference. Importantly, it's worth mentioning that python-numpy file will be made available.

\begin{table*}[ht]
  \centering
  \caption{The sensory apparatus employed in this dataset and accompanying technical specifications.}
  \label{tab:sensorspecificaions}
   \renewcommand{\arraystretch}{1.5}
  \begin{tabular}{llllc}
    \toprule
    \hline
    \textbf{Sensor and Data Characteristics} & \textbf{Model} & \textbf{Topic Name} & \textbf{Message Type} & \textbf{Rate (Hz)} \\
    \midrule
    Camera: EM 380-800nm & Pixel-XYZ AR135-130T400 & /usb\_cam/image\_raw & sensor\_msgs/Image & 30 \\
    \hline
    Conic LIDAR: EM 905nm & Livox Avia & /livox\_avia/lidar & livox\_ros\_driver/CustomMsg  & 10 \\
    \hline
    Peripheral LIDAR: EM 905nm & Livox Mid360 & /livox\_mid360/lidar & livox\_ros\_driver2/CustomMsg & 10 \\
   \hline
    \multirow{4}{*}{Audio Array: Acoustic 0-44.1 kHz } & \multirow{4}{*}{Hikvision DS-VM1} 
    & /audio1/audio & audio\_common\_msgs/AudioData & \multirow{4}{*}{41.8} \\
    && /audio2/audio & audio\_common\_msgs/AudioData & \\
    && /audio3/audio & audio\_common\_msgs/AudioData & \\
    && /audio4/audio & audio\_common\_msgs/AudioData & \\
   \hline
    \multirow{3}{*}{MMwave Radar: EM 77GHz} & \multirow{3}{*}{Oculii Eagle ETH04} 
    & /radar\_enhanced\_pcl & sensor\_msgs/PointCloud & \multirow{3}{*}{15} \\
    && /radar\_pcl & sensor\_msgs/PointCloud & \\
    && /radar\_trk & sensor\_msgs/PointCloud & \\
   \hline
    \bottomrule
    
  \end{tabular}
  \vspace{-1em}
\end{table*}

\section{Sensor Calibration}
This sensor array encompasses a diverse range of sensing modalities, each with its unique calibration challenges. To address these challenges effectively, we adopt a divide-and-conquer approach.

For stereo calibrations, we leverage Matlab calibration tools\cite{zhang2000flexible} to precisely calibrate the intrinsic and extrinsic parameters between the two cameras. When it comes to camera-to-LIDAR calibration, we employ a targetless calibration method\cite{yuan2021pixel}, ensuring accurate alignment.

Calibrating audio and mmWave Radar sensors present greater complexity due to their limited correspondences with other modalities. In these scenarios, we resort to CAD drawings to establish reference points, specifically aligning them with the top-centered Livox Mid360 LIDAR. This approach becomes essential, considering the inherent intricacies of these sensors and their incompatibility with traditional calibration techniques.

\section{Evaluation and Benchmark}
This section presents baseline benchmarking of various image and audio-based 2D detection and 3D estimation methods. Due to page limitations, we will not cover the LIDAR or Radar-based model here, but their performance and source code will be available on the dataset webpage.

We gathered over 1700 seconds of multi-modality in rosbag format, which was subsequently divided into 50 smaller sequences. Each of these sequences includes ample visual, Lidar, audio, and radar data for identification purposes. Within these 50 smaller sequences, we allocated 60\% of the data for the training set, 20\% for the testing set, and the remaining 20\% for the validation set.

Each model exhibits slight variations in training parameters, particularly in learning rates. However, it's worth noting that all models share a common batch size of 8 during training. We assess the 2D detection benchmark using mean average precision (mAP) and frame per second (FPS) measurements to evaluate the performance of these popular methods on our dataset. The results are presented in Table \ref{tab:2d_detection}. 

\begin{table}[h]
\centering
\caption{2D Object detection accuracy and running speed performance for all types of drones.}
\renewcommand{\arraystretch}{1.5}
\setlength{\tabcolsep}{14pt}
\label{tab:2d_detection}
\begin{tabular}{lcc}
\hline
Model & mAP & FPS\\
\hline

Yolov5\cite{glenn_jocher_2020_4154370}     & 84.85 &  \textbf{30.2} \\
\hline
YoloX\cite{ge2021yolox}      &  \textbf{85.9 } & 23.3 \\
\hline
SSD\cite{liu2016ssd}         & 71.7  & 22.8\\
\hline
Centernet\cite{duan2019centernet}   & 81.9  & 28.8 \\
\hline
M2det\cite{zhao2019m2det}       & 76.0  & 21.1 \\
\hline
\hline
\end{tabular}
\vspace{-1em}
\end{table}

\begin{figure}[t]
\centering
\includegraphics[width=8.6cm]{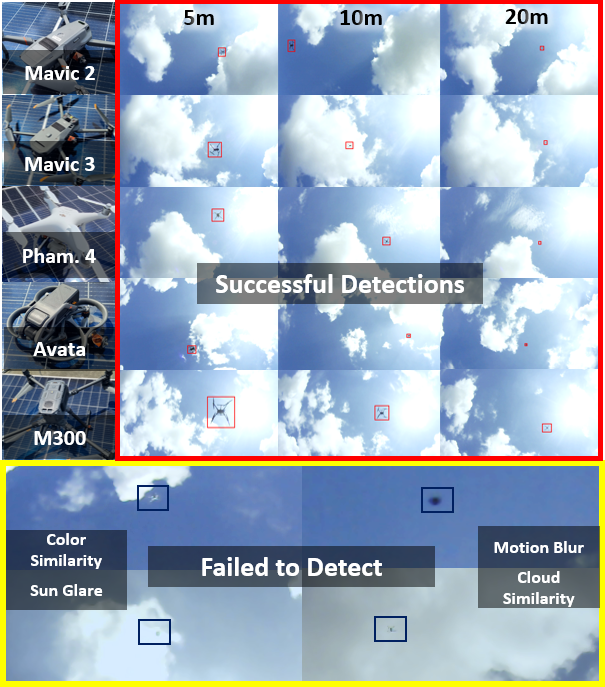}
\caption{The visual detection results encompass both successful and unsuccessful cases when employing YoloX. Drones that are white, smaller in size, or located at a greater distance are less likely to be detected.}
\label{fig:visual_detection_result}
\vspace{-2em}
\end{figure}

\begin{figure}[t]
\centering
\includegraphics[width=7.8cm]{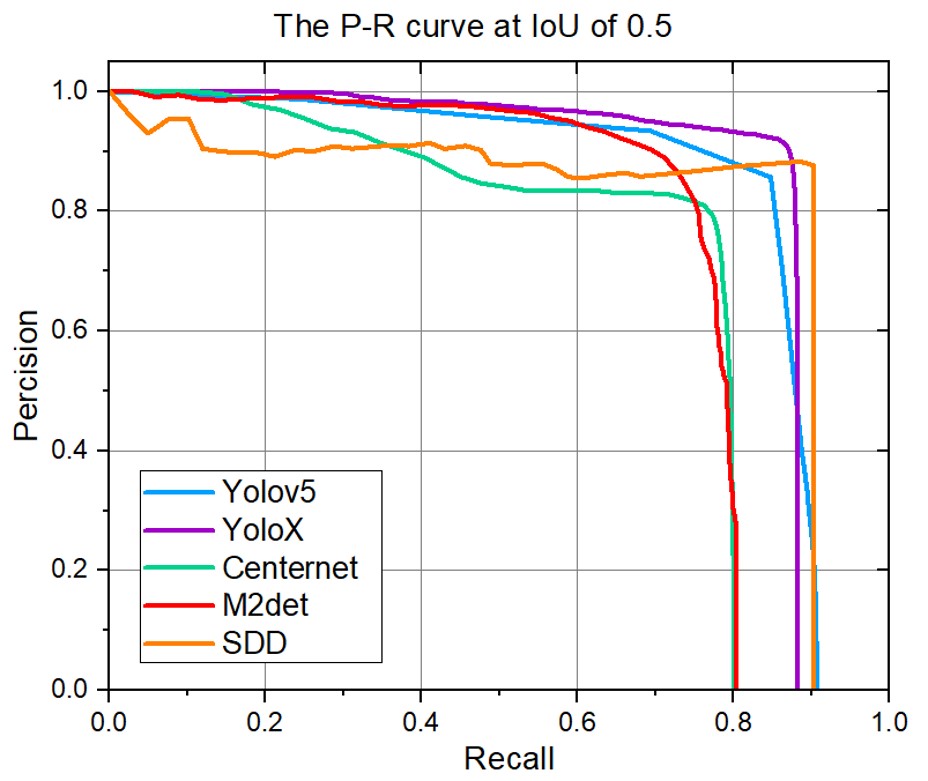}
\caption{Precision-recall curve of visual detection result.}
\label{fig:precisionandrecall}
\end{figure}

Currently, no models have been fine-tuned or pruned for improved performance. Among the popular networks, the Yolo series can mostly outperform other methods, as shown in Fig. \ref{fig:precisionandrecall}.  Models like Centernet excel in detecting other types of vehicles, such as cars or trucks, but experience a significant drop in performance when it comes to UAV prediction. This is mainly because UAV targets are considerably smaller than other vehicles, as illustrated in Fig.\ref{fig:visual_detection_result}.

In the context of 3D position estimations, we made adaptations to several well-known networks to determine the 3D positions of UAV targets based on either audio or visual inputs. Similar to the 2D detection scenario, the learning rate varies between cases, but we maintain a consistent batch size of 8.
In our assessment of 3D position estimations, we employ the term relative position error $e$ to gauge the disparities between model outputs and ground truth.
\begin{table}[h]
\centering
\caption{3D position estimations accuracy and all types of drones for error in x, y, z and average relative distance error. Note, $e_r$ is not he L2 norm of error in x,y and z.}
\label{tab:3d_ape}
\renewcommand{\arraystretch}{1.5}
\begin{tabular}{lccccc}
\hline
Modality & Network & $e_x$ & $e_y$ & $e_z$ & $e_r$\\
\hline
\multirow{5}{*}{Visual} & Resnet50\cite{he2016deep}   &  \textbf{0.12} & 0.32 & 0.30 & 0.55 \\
 & Resnet101\cite{he2016deep}   & 0.177 & 0.35 & 0.27 &  \textbf{0.54} \\
 & Vgg16\cite{SimonyanZ14a}       & 0.16 & 0.40 & 0.23 & 0.55 \\
 & Vgg19\cite{SimonyanZ14a}       & 0.13 &  \textbf{0.29} & 0.38 & 0.57 \\
 & Darknet\cite{bochkovskiy2020yolov4}     & 0.18 & 0.44 &  \textbf{0.14} &  \textbf{0.54} \\
\hline
\multirow{2}{*}{Audio} & VorasNet\cite{vora2023dronechase}    & \textbf{0.54} & \textbf{1.59} & \textbf{1.51} & \textbf{2.64} \\
 & Audio Transformer\cite{park2021many}   & 0.71 & 2.07 & 1.99 & 3.43  \\
\hline
\end{tabular}
\vspace{-1em}
\end{table}

The comprehensive performance evaluation for 3D position estimation is summarized in Table \ref{tab:3d_ape}. All visual models exhibit a consistent error of slightly over 0.5 meter, indicating their capability to accurately estimate object size and provide reasonable object distance and position estimates. However, the audio-based approach faces significant challenges due to the substantial impact of background machinery noise, resulting in a 2.6-meter error. 
It's important to highlight that there is a limited availability of compatible open-source algorithms in this category, which has resulted in a narrower scope of research in this area.

\section{Issues and Challenges}
In the course of our research, we encountered a set of substantial challenges and limitations that must be acknowledged.
\subsection{Limited Geographic Coverage}
Our data collection efforts were significantly constrained by the restricted geographical areas available for drone flights. Singapore regulatory framework mandates obtaining permits from landowners, and nearly 70\% of locations are designated as no-fly zones due to their proximity to airports and airbases. Consequently, our ability to conduct drone experiments across diverse environments was severely limited. To ensure compliance with local privacy laws, which prohibit the capture of images featuring individuals or vehicles, we conducted most of our experiments from rooftops.
\subsection{Sensor Synchronization}
We sought to achieve synchronization among all sensors to improve the quality of our training data. However, the majority of sensors lack inherent synchronization capabilities. Our budget-conscious camera, while cost-effective, lack the ability to be externally triggered. Furthermore, the audio data was sampled at a rate of 41.8kHz, which significantly deviated from the sampling rates of other modalities. As a result, achieving synchronization across all modalities without incurring substantial hardware costs posed a formidable challenge.
\subsection{Limited Drone Variability}
Our dataset comprises data from only a limited selection of publicly available drone models. Considering the extensive diversity of drones available, our data collection efforts represent a mere fraction of the UAV types in existence. The considerable variation in drone designs further complicates our ability to compile comprehensive data across all drone categories.
\subsection{Ground Truth Rate and Missing Data}
The ground truth data generated by Leica is captured at a rate of only 5Hz. While this rate serves the needs of many applications, it may not be sufficient for certain research scenarios that demand higher temporal precision. Additionally, occasional gaps in the ground truth data may occur when drones execute sharp turns or when Leica struggles to track all sides of the UAV simultaneously.

Notwithstanding these challenges and limitations, our research contributes valuable insights and datasets to the field of drone detection, tracking, classification and trajectory estimation.

\section{Conclusion}
In summary, the MMAUD dataset represents a significant advancement in addressing the challenges posed by small UAVs. It focuses on aerial detection, UAV-type classification, and trajectory estimation, bridging a critical gap in current aerial threat detection methodologies. MMAUD leverages diverse sensory inputs, including stereo vision, Lidars, Radars, and audio arrays, offering a unique and practical perspective.

An essential highlight of MMAUD is its reliance on Leica-generated ground truth, recognized for its remarkable accuracy in outdoor practical scenarios which is a feature unmatched by other datasets. While we acknowledge some limitations, such as regulatory constraints on data collection and sensor synchronization challenges, MMAUD remains a valuable resource for developing precise anti-UAV solutions.

\bibliographystyle{IEEEtran}
\bibliography{mybib}


\end{document}